\documentclass{article}

\usepackage[preprint]{neurips_2021}
\usepackage{natbib}

\usepackage[utf8]{inputenc} 
\usepackage[T1]{fontenc}    
\usepackage{hyperref}       
\usepackage{url}            
\usepackage{booktabs}       
\usepackage{amsfonts}       
\usepackage{nicefrac}       
\usepackage{microtype}      
\usepackage{xcolor}         

\title{SparseChem: Fast and accurate machine learning model for small molecules}

\author{%
  (Adam Arany,~ Jaak Simm)\thanks{Contributed equally as first authors.},~ Martijn Oldenhof and Yves Moreau \\
  ESAT STADIUS, KU Leuven, Leuven, 3000, Belgium\\
  \texttt{\{jaak.simm,adam.arany,}\\\texttt{yves.moreau,martijn.oldenhof\}@esat.kuleuven.be}\\
}

\begin{document}

\maketitle

\begin{abstract}
  \textbf{Summary:} SparseChem provides fast and accurate machine learning models for biochemical applications. Especially, the package supports very high-dimensional sparse inputs, \emph{e.g.}, millions of features and millions of compounds. It is possible to train classification, regression and censored regression models, or combination of them from command line. Additionally, the library can be accessed directly from Python.\\
\textbf{Availability and implementation:} Source code and documentation is freely available under MIT License on GitHub (\href{https://github.com/melloddy/SparseChem}{https://github.com/melloddy/SparseChem}). 
\end{abstract}

\section{Introduction}
Training a machine learning model for a specific task when high quality data is scarce is a challenging problem. The paradigm of multi-task learning~\citep{caruana1997multitask} enables the use of training data from related tasks to jointly train a machine learning model. In several applications~\citep{ruder2017overview} multi-task learning has been successful and allows a machine learning model to be more accurate in predicting multiple tasks compared to using multiple machine learning models predicting one task at a time. Also in the field of drug discovery multi-task learning has gained importance~\citep{simoes2018transfer,dahl2014multi} in recent years for quantitative structure–activity relationship (QSAR) models. 


Prediction of drug bioactivity and toxicity is crucial for development of new drugs.
Modern chemical descriptions, such as ECFP \citep{rogers2010extended}, used with machine learning methods are the \emph{de facto} standard for such ligand-based activity modelling \citep{Macau, deeptox}.
In the paper we present a flexible multi-task machine learning tool \textbf{SparseChem} that allows an easy and scalable way to train neural networks with \emph{high-dimensional} sparse input features like ECFP.
SparseChem was built with industry scale data sets in mind, and for example, can complete a full training run of 420,000 compounds and 3,500 tasks in 2 minutes on a modern GPU, \emph{e.g.}, NVIDIA V100.

Next we outline the main features of SparseChem.
\begin{enumerate}
    \item \textbf{Supports high dimensional input space.} SparseChem uses \emph{sparse} neural network layer as its input layer and thus can handle even million dimensional (sparse) inputs. For example, full ECFP descriptor can contain millions of features.
    \item \textbf{Massive multi-task setups.} By using a sparse training loss, only defined on the output elements that are observed, SparseChem can support high number of tasks, e.g., 30,000 tasks.
    \item \textbf{Internal mini-batching} allows to use larger effective mini-batch size than what would fit into the GPU memory.
    \item \textbf{Support for classification, regression and censored regression}. SparseChem implements loss functions for several data setups, logistic loss for classification, MSE for regression and the censored MSE for censored regression.
    \item \textbf{Hybrid models} containing both classification and (censored) regression in the same network. This allows for quicker training and can improve accuracy in some multi-task setups.
    \item \textbf{Individually specified weights for each task}. SparseChem allows the user to provide a list of tasks weights to fine-tune the training to the specific problem at hand, \emph{e.g.}, treating some tasks as auxiliary tasks with lower weights.
    \item \textbf{Support for many metrics} used for classification and regression out of the box, computed per each task, such AUC-PR, AUC-ROC, F1, Kappa for classification and correlation, R$^2$, RMSE for regression.
    \item \textbf{Support for CPU and GPU hardware}. Both training and inference can be run on either CPU or GPU hardware.
\end{enumerate}

\section{Supported training modes}
As mentioned before in addition to standard training losses, such as binary cross-entropy for classification and mean squared error for regression, SparseChem also supports censored regression. In that mode each data point can either have no censoring (\emph{i.e.}, standard least squares loss), \emph{upper} censoring or \emph{lower} censoring.

Both in classification and in a non-censored setting, the user should prepare a (sparse) input matrix $\mathbf{X}$ and (sparse) output matrix $\mathbf{Y}$ of the target values. 
In the case of censored regression, the user additionally has to provide a \textbf{censoring mask} matrix $\mathbf{C}$ whose entries correspond one to one to the entries of $\mathbf{Y}$:
\begin{enumerate}
    \item \textbf{Upper censoring:} entry $\mathbf{C}_{ij}$ should be $+1$,
    \item \textbf{Lower censoring:} entry $\mathbf{C}_{ij}$ should be $-1$,
    \item \textbf{No censoring:} entry $\mathbf{C}_{ij}$ should be $0$.
\end{enumerate}

For the upper censoring the loss is a one-sided square loss:
\begin{equation}
    L_\mathrm{up} = \max(\mathbf{Y}_{ij} - \widehat{\mathbf{Y}}_{ij}, 0 )^2,
\end{equation}
where $\mathbf{Y}_{ij}$ is the upper-censored observed value and $\widehat{\mathbf{Y}}_{ij}$ is the prediction.
If the prediction $\widehat{\mathbf{Y}}_{ij}$ is larger than the censored value the loss becomes zero.

Similarly, the lower censoring loss is
\begin{equation}
    L_\mathrm{low} = \min(\mathbf{Y}_{ij} - \widehat{\mathbf{Y}}_{ij}, 0 )^2.
\end{equation}

\section{Software}
The SparseChem package provides command line interface for both training and inference. Additionally, the sparse linear layer introduced by SparseChem is also available as PyTorch Module.

In the following sections we demonstrate how to use SparseChem for classification and regression.

The execution times and predictive performances were measured on CheEMBL 29 \citep{chembl} filtered according to \citep{simm2021expressive}, Appendix C. The ECFP radius 3 fingerprints were calculated using RDKit 2019.03.3. The dataset contains 424k compounds, 888 assays, over all 666299 measurements, resulting in a sparsity of 0.18 \% . The 5-fold cross-validation was created on chemical clusters as described in \citet{imaging}.
The execution times was measured on a single Nvidia TITAN Xp GPU. All training was run for 20 epochs.

\subsection{Classification example}

To start the training on a classification data set we can use two Numpy files containing the training features (X.npy) and labels (Y.npy) as a sparse matrix:

\begin{verbatim}
 python -m sparsechem.train --x X.npy \ 
   --y_class Y.npy --folding folds.npy \
   --weight_decay <wd> --hidden <h> \
   --dropout_trunk <do> 
\end{verbatim}

The arguments setting hyper-parameters will be omitted in the further examples. Please consult the documentation for more details.
We observed AUC-ROC = 0.794, AUC-PR = 0.717 with execution speed of 12 epochs/min. The corresponding hyper-parameters are 2000 hidden neurons, 0.6 of dropout and no weight decay.

\subsection{Regression example}
\begin{verbatim}
 python -m sparsechem.train --x X.npy \ 
 --y_regr Y.npy --folding folds.npy \
 --standardize_regression 1 [...]
\end{verbatim}

The argument normalize\_regression enable standardization of the data set.
We observed R-squared = 0.358, Pearson correlation = 0.620 with execution speed of 15 epochs/min. The corresponding hyper-parameters are 1000 hidden neurons, 0.7 of dropout and 6e-4 of weight decay.

\subsection{Censored regression example}

\begin{verbatim}
 python -m sparsechem.train --x X.npy \
 --y_regr Y.npy --y_censor C.npy \
   --folding folds.npy 
   --normalize_regression 1 [...]
\end{verbatim}
We observed R-squared = 0.356, Pearson correlation = 0.621 with execution speed: 11 epochs/min. The hyper-parameters are identical to the regression case. Note that in this data set using the censored values does not result in significant difference in predictive performance.

\subsection{Prediction on new data}

To generate predictions on a new set of instances, we can use the following command:

\begin{verbatim}
 python -m sparsechem.predict --x Xnovel.npy \
 --conf model.json --model model.pt \
   --outprefix out
\end{verbatim}

The predictions will be saved to Numpy data files (.npy) corresponding to the specified name prefix.

\section*{Acknowledgements}

AA, JS, MO and YM are funded by (1) Research Council KU Leuven: C14/18/092 SymBioSys3; (2) CELSA/21/019, (3) the EU/EFPIA Innovative Medicines Initiative 2 Joint Undertaking (MELLODDY grant n°831472), (4) Flemish Government (ELIXIR Belgium, IWT: PhD grants) and (5) Impulsfonds AI: VR 2019 2203 DOC.0318/1QUATER Kenniscentrum Data en Maatschappij. Computational resources and services used in this work were partly provided by the VSC (Flemish Supercomputer Center), funded by the Research Foundation - Flanders (FWO) and the Flemish Government – department EWI. We also gratefully acknowledge the support of NVIDIA Corporation with the donation of the Titan Xp GPU used for this research.

We thank MELLODDY consortium members for feedback and suggestions. Especially, Rene Mueller from NVIDIA.



\bibliographystyle{plainnat}
\bibliography{document}
\end{document}